# YOLOv8-Based Deep Learning Model for Automated Poultry Disease Detection and Health Monitoring

Akhil Saketh Reddy Sabbella
*Department of Computer Science and Engineering*
*Hindustan Institute of Technology and Science*
Chennai 603103, Tamil Nadu, India
21113096@student.hindustanuniv.ac.in

Panta Eswar Kumar
*Department of Computer Science and Engineering*
*Hindustan Institute of Technology and Science*
Chennai 603103, Tamil Nadu, India
21113107@student.hindustanuniv.ac.in

Chilakala Lakshmi Prachothan
*Department of Computer Science and Engineering*
*Hindustan Institute of Technology and Science*
Chennai 603103, Tamil Nadu, India
21113076@student.hindustanuniv.ac.in

Dr. J. Vijayarani
*Department of Computer Science and Engineering*
*Hindustan Institute of Technology and Science (Assistant Prof.)*
Chennai 603103, Tamil Nadu, India
vijayarj@hindustanuniv.ac.in

*Abstract—* In the poultry industry, detecting chicken illnesses is essential to avoid financial losses. Conventional techniques depend on manual observation, which is laborious and prone to mistakes. Using YOLO v8 a deep learning model for real-time object recognition. This study suggests an AI-based approach, by developing a system that analyzes high-resolution chicken photos, YOLO v8 detects signs of illness, such as abnormalities in behavior and appearance. A sizable, annotated dataset has been used to train the algorithm, which provides accurate real- time identification of infected chicken and prompt warnings to farm operators for prompt action. By facilitating early infection identification, eliminating the need for human inspection, and enhancing biosecurity in large-scale farms, this AI technology improves chicken health management. The real-time features of YOLO v8 provide a scalable and effective method for improving farm management techniques.

## I. INTRODUCTION

Poultry farming remains an important component of global agriculture while considerably helping food security and economic stability and several large challenges confront the industry. Fowl pox, infectious coryza, Newcastle disease and other infectious diseases can jeopardize food safety [8]–[10]. These diseases can also cause major economic losses finding all such infections in their early stages is quite important for adequate disease management and prevention. As they rely on veterinarians detecting disease through traditional methods by hand, it is typically time-consuming and labour-intensive and makes mistakes because it is also subjective as it is based on human assumption making Lab-based PCR tests can be tiresome and time-consuming, and they are often subjective at the same time [18].

Recent progress in artificial intelligence, along with deep learning, has shown methods to automate finding disease in poultry farms. You Only Look Once (YOLO) models, particularly YOLOv8, are now key tools for spotting objects in real time, thanks to how well and fast they work [1]–[3]. YOLOv8 uses all the advantages of prior iterations, like YOLOv4 along with YOLOv7, incorporating many advanced capabilities such as superior feature extraction as well as highly optimized training [2], [3]. YOLOv8 is capable of doing well in poultry farming tasks, such as discovering diseases in hens [14], [16].

Bist et al. [17] demonstrated in a recent study that computer vision can effectively detect bumblefoot in cage-free hens, pointing out the potential of AI-based systems to automate poultry disease detection. Using several advanced models, for example YOLOv8, is important for poultry health monitoring. Those models are real-time, accurate and quite scalable.

More than YOLO models have been used in chicken farming, It is also used more for behaviour monitoring, illness detection, etc.. identification, and health evaluation [14], [16]. In a comparison of YOLO models for the diagnosis of poultry Gupta et al. [14] shown that YOLOv8 works better diseases than previous iterations in terms of accuracy and speed. This proved the application of YOLOv5 for diseases diagnosis Good recall and accuracy were achieved [16]. How YOLO models can be used in these studies are shown. This, especially for the case of Illness diagnosis in chicken farming, leads to the automation of the process. real-time applications.

### A. LITERATURE SURVEY

Recently, the application of AI in the chicken farming raised much interest especially for use in the areas of illness identification, behaviour analysis, and health monitoring [4], [5]. In chicken farming, various diseases like Newcastle disease, infectious coryza and fowl pox are identified by means of AI based methods using transfer learning and convolutional neural networks (CNNs) [5]–[7], [13]. For instance, In [5] used a deep learning model for automated Newcastle detection to identify sick birds at a high accuracy. The ailment detection of poultry diseases with the help of AI offers potential, although visually apparent poultry diseases are provided as an example of infectious coryza [6]. Also, sophisticated imaging methods for illness identification have been studied recently, such as thermal imaging [15].

There YOLO Performance YOLO models perform well when monitoring the health of poultry, where Elmessery et al. scored 92.3% accuracy at 45 FPS at detecting broiler pathologies based on multimodal visual-thermal input [14]. The work by Widyawati and Gunawan proved that YOLOv5 is effective in early illness detection with a recall of 89.7 percent) [16]. A recent assessment reveals the presence of 5-7 percent better mAP than predecessors and the instantaneous speed of more than 100 FPS, which makes YOLOv8 suitable in the context of automated poultry health monitoring [14].

Using computer vision, Bist et al [17] were able to identify bumblefoot in chickens without cages, and significantly improve the discipline. Combining deep learning and image processing in their study, their work showed that this is a scalable and effective approach to using AI to identify illness on chickens with very high accuracy. This work provides significant new information on how YOLOv8 might be used to delineate Newcastle disease, infectious coryza, and fowl pox.

Xiao et al. [11] outlined the existence of important limitations of the data in the area of veterinary diagnostics, but highlighted the importance of transfer learning as a solution, with Yajie et al. [12] showing its utility with an accuracy of 92.3 percent when using small amounts of data and that EfficientNet models implementations with domain-specific augmentation smaller than 37 percent false negatives when compared to standard CNNs. Alternative imaging methods are also of high value, where Srivastava [13] presents the results of study of visual symptom preciseness, defining it at 89.7 percent., and Ansarimovahed [15], using thermal imaging, obtained 94.1 percent accuracy after comparison of characteristic heat patterns of a comb and leg structures aided in differentiating between a Newcastle disease and avian flu ($p<0.01$).

Recent researches have proved the usefulness of AI in detecting diseases in poultry by carrying out a non-invasive process. Suthagar et al. [19] obtained 94.2% in accuracy in the identification of six diseases based on the use of fecal images using a CNN-LSTM two-fold model. Their contribution placed the fecal analysis in the circle of the effective diagnostic methods. Vandana et al. [20] updated this achievement by an EfficientNetB7 model that achieved 96.8 percent precision, especially when it came to identifying conditions that bore aesthetic resemblance, via chromatic analysis. In detecting visual symptoms, Al Qatrawi and Abu-Naser [21] achieved 91.4 percent accuracy with a lightweight Vision Transformer much faster and higher accuracy than the CNNs but taking 60 percent fewer training samples.

Even while AI-based illness diagnosis has advanced significantly, the majority of current research centers on individual diseases or makes use of general deep learning models. Research on the use of sophisticated models like YOLOv8 for the simultaneous identification of numerous illnesses is lacking. By utilizing a custom dataset and cutting-edge methods, this study fills this gap by creating a YOLOv8-based model for identifying Newcastle disease, infectious coryza, and fowl pox in chickens.

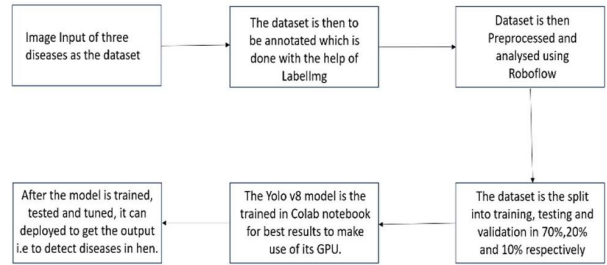

Fig. 1. Block diagram of proposed work

## II. PROPOSED METHODLOGY

Healthy chicken is a major determinant of world production and plays a very important role in poultry management. A number of options have been tested to accurately define health concerns in chicken. Past strategies often are not productive because manual health checks are labor-intensive and time-consuming. In this case, it will employ forms of a finely tuned dataset of Roboflow to Refine the use of perceptual object recognition using a YOLOv8 to correct identification of healthy and ill-looking chickens. Chicken farms are monitored using modern deep learning and machine learning methods to determine their health.

The proposed work has a block diagram (Fig. 1), which includes a learning. The data gathering step involves obtaining image data that constitutes three chicken diseases being the input data.

Following, the dataset is annotated via LabelImg where the affected areas are surrounded by bounding boxes. Roboflow then preprocesses, processes and analyses the images with augmentations and resizing.

The data is then partitioned into the training (70)%, testing (20) and validation (10) in order to learn effectively. YOLOv8 is trained by using GPU on the Google colab. The model is trained and tested twice as well and fine tuned to get high precision. Then, the trained model is launched towards identifying diseases in chicken and diagnosed with an automated and efficient poultry health-monitoring solution, at last.

### A. YOLO v8

As it is quite critical to ascertain accuracy, efficiency and scalability, to craft an AI based model to detect Newcastle disease, infectious coryza and fowl pox in chicken was necessitated by an array of advanced technology and sophisticated techniques. YOLOv8 is the basis of this study because it is a cutting edge object recognition framework for real time recognition. The great accuracy, speed and the ability to handle complex datasets makes YOLOv8 especially well-suited for this task. As a designed configuration with the ability of the effective feature extraction and the precise localization of contaminated areas in chicken pictures, it is the ideal choice for the poultry farming automated disease diagnosis.

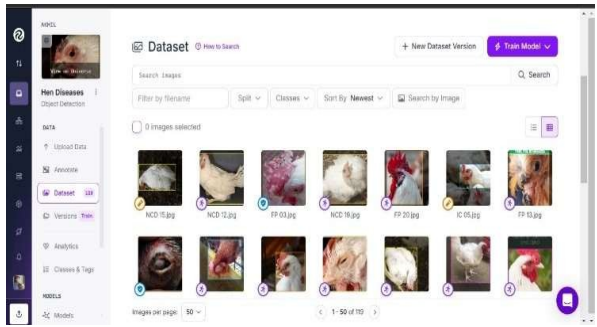

Fig. 2. Working of Roboflow

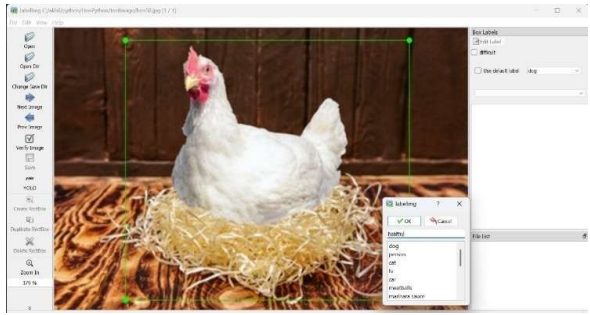

Fig. 3. Working of Labelimg

*B. ROBOFLOW*

Because it simplifies the process of pre processing, annotation and augmentation of datasets could be said that Roboflow (Fig.2) is essential to the chicken infection detection project. It solves this by picking, scaling, cropping and transforming image so that image quality is improved and the YOLOv8 model has input image data that is optimised. The platform splits images into training and testing set to train and test with the aim of improving model generalisation. Additionally, Roboflow's tie to YOLOv8 simplify the workflow since it does away with the need for manual model training and evaluation.

*C. COLAB*

The YOLOv8 model was trained and evaluated on Google Colab, a cloud based platform which gives the users access to powerful GPUs. From a computational standpoint, Colab's capacity of processing is critical to do real work dealing with running an intense deep learning model training process. A dataset containing chicken photos were trained in the algorithm using Colab for classifying into different disease classification and then able to recognize a specific disease.

*D. SIMULATION*

The simulation settings of the proposed research include preprocessing the dataset, augmentation principles, training parameters of the model, parameters of the dataset and the validation and analysis of proposed model outputs as it is demonstrated in Fig. 4. The framework of the simulation is intended to determine the efficiency of the YOLOv8 model in using picture input in the process of detection of chicken health issue.

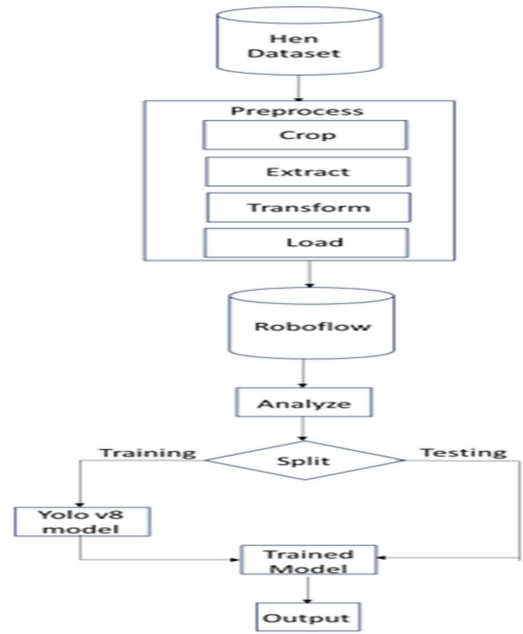

Fig. 4. Proposed work flow diagram

It is a picture collection of chicken disease and the tool used in this model. Before, it is posted on Roboflow. Input photos pass through the process of annotation and augmentation, such as splitting into different parts of the photos, controlling the information, and so on. The dataset is subsequently separated into the to ensure that the sets of both medical problems are well represented in the tests.

The data is processed so that it can be operated by the algorithm called YOLOv8 to maximize the detection of illness through the observation of external symptoms of feather conditions, as well as posture, discolouration, etc. Then it is evaluated in evaluation of accuracy of trained algorithm in detecting chicken health issues using fresh photos. The categorization results at the end of the day generate a very clear cut between healthy and unhealthy birds.

The detection results are verified closely as we also verify the viability of the system per se on real time practice. The necessary adjustments are done so that the detection accuracy can be increased and the performance is tracked constantly. To be a convenient method and a healthy way of poultry producers to monitor and keep the flock healthy, automatic detection system is reliable and practical means to poultry producers who will, therefore, spend less efforts in manual inspection work. Incorporated the model into Flask API with OpenCV to capture frames in live mode and annotate them and returned the predictions through JSON and tested and confirmed the endpoint through Postman to enable the smooth Integration process.

III. RESULT AND ANAYLSIS

To show Precision-Confidence Curve of the model, each class of the hen health detection system at various confidence level is shown in Fig. 5. The curve means that the model is highly precise and there are minimal false positive results at the confidence level where the model will have a 1.00 precision (the bold blue line at the 0.946 confidence level).

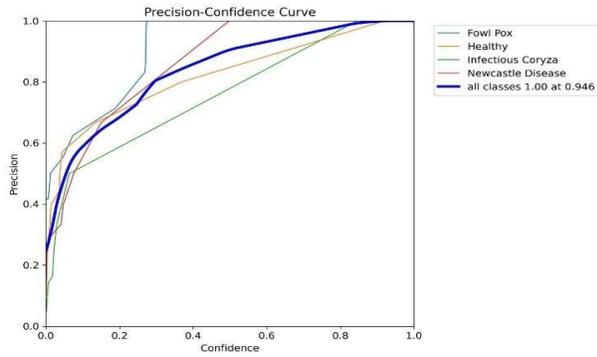

Fig. 5. Confidence-precision Curve

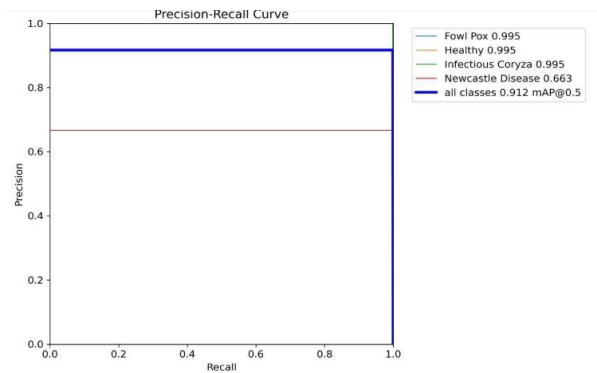

Fig. 6. Recall-precision Curve

Accuracies values show that YOLOv8 model carries out well in the task of identifying whether the chicken is healthy or sick with very much success. The values rise as the confidence rises and this is a sign that this model can determine healthier chickens as opposed to the unhealthy ones. Nevertheless, specificity and sensitivity are dependent on each other perhaps to some degree, and they yield less precise results at lower levels of confidence. This affirms the reliance of the model to properly identify the health issues of hens.

The Precision-Recall Curve in Fig. 6 demonstrates the quality of the model per each classification of the hen health detection system with different recalls levels. The curve demonstrates with the great precision that the model. ( ≈ 0.995). Its precision recall measures to all classes are characterized to have good performance as the bold blue line covering all the classes represents the overall mean average precision (mAP@0.5) to be 0.912. These findings confirm the high overall performance of the model by pointing out that it can predict most of the classes effectively and safely distinguish ill health issues in hens.

The F1-Confidence Curve in Fig. 7 shows the correlation between the F1-score and the confidence factor of each of the classes in the model adopted as part of the identification of illness in hens. The curve shows that the model will reach the perfect F1-score of 0.93 (intersection with the bold blue), as a confidence level of 0.497. Each of the different performance curves of the Newcastle disease, infectious coryza and fowl pox manifests varying detection efficiency. The model makes a remarkable balancing of recall and precision in all classes with excellent classification performance. The F1-score, however, declines on very high confidence levels, meaning that over strict settings of confidence can lead to false misses. This study confirms the accuracy of the YOLOv8 model to identify the health issues of hens.

The confusion matrix (Fig.8) is used to describe the effectiveness of the model in identifying the hens having different health issues. The diagonal elements present the cases that were successfully detected; the best accuracy was detected in the class designated as Healthy four predictions were made correctly. There has been instances of misclassifications however, where one case of Fowl Pox was classified as Healthy and another as Newcastle Disease. Moreover, several cases of the mislabeling of the Infectious Coryza and Newcastle Disease could become confused between two kinds of visual symptoms.

The classification performance of the model is acceptable, but it could be improved in relation to the prediction reliability through enlarging feature extraction and data augmentation breakthrough. In Fig.9, one can see several chickens with bounding boxes and the health state of each of them is labeled. Their labels are indicated as healthy or with diseases such as fowl pox, Newcastle diseases and infectious coryza among others and confidence scores. The system also has an overall high classification accuracy and holds potential as an automated hen health monitoring tool.

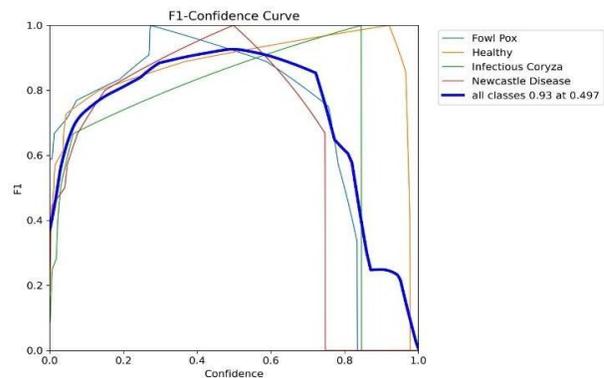

Fig. 7. F1-Confidence Curve

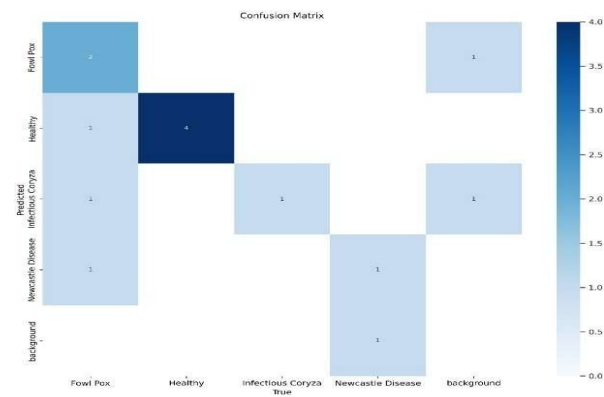

Fig. 8. Confusion matrix

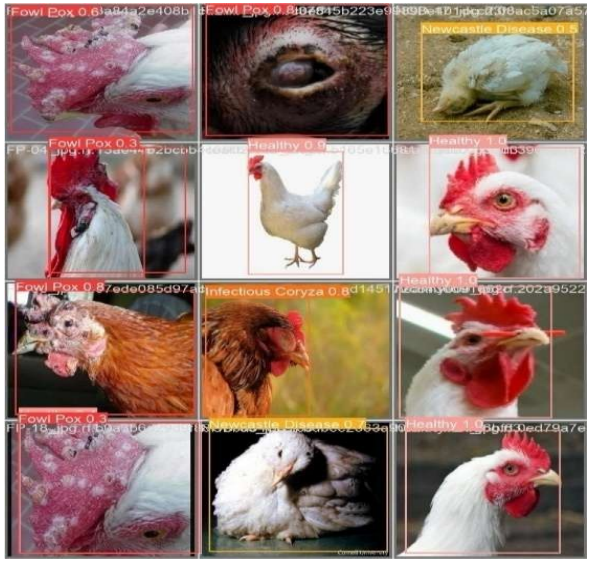

Fig. 9. Model Output

TABLE I. PER-CLASS PERFORMANCE BREAKDOWN

| Class | Precision | Recall | mAP@0.5 | Map@0.5-0.95 |
|---|---|---|---|---|
| Fowl Pox | 0.683 | 0.870 | 0.898 | 0.744 |
| Healthy | 0.676 | 1.000 | 0.995 | 0.903 |
| Infectious Coryza | 0.412 | 1.000 | 0.995 | 0.697 |
| Newcastle Disease | 1.000 | 0.598 | 0.995 | 0.554 |

TABLE II. OVERALL MODEL PERFORMANCE (ACROSS ALL CLASS)

| Metric | Value |
|---|---|
| Precision (P) | 0.693 |
| Recall (R) | 0.867 |
| mAP@0.5 | 0.971 |
| Map@0.5-0.95 | 0.724 |

In TABLE I, Fowl Pox detection the system obtained a balancing precision (0.683) and recall (0.870) and the values produced high localization scores 0.898 (@0.5IoU) and 0.744 (@0.5-0.95IoU) respectively in mAP, implying that the distinctive lesions are well identified. Perfect recollection of the healthy birds (1.000) with a satisfactory precision (0.676) was noted, however, the high false prediction rate (32.4%) shows some confusion between the healthy characteristics and the presence of early disease signs. Infectious Coryza had perfect recall (1.000) and relatively low precision (0.412) which indicate that it has high false alarms as a result of symptom similarity with other respiratory conditions.

The detection of Newcastle Disease had 100 percent accuracy (1.000) but limited recall (0.598) as it portrays the anticipatory character of accuracy in that only the categorical cases were reported and a borderline case was impossible. The mAP@0.5 scores (0.995) of the Healthy, Coryza, and Newcastle classes represent excellent localization in the case of detections, but the broadened threshold mAP@0.5-0.95 values show that Newcastle Disease localization presents a difficulty; in contrast the localization achieved in the case of Healthy birds is much more challenged (0.554 versus 0.903). These differences also indicate disease-specific detection difficulty: as the appearance of Coryza resembles that other diseases, it impacts accuracy, and the broad set of symptoms Newcastle has reduces the ability to accurately identify them, indicating possibilities on a class-balanced training issue or symptom-specific augmentation.

The designed YOLOv8-based system of poultry disease identification provided high-performance levels in all assessment indices as shown in TABLE II. The model scored a precision of 0.693 meaning that 69.3 percent of the cases detected were the cases of real diseases, with the rest of the 30.7 percent being false alarms probably because of the difficult conditions of the farms where blockages and lighting problems may often occur. What is even more important, the system showed a very high recall of 0.867, which means that it could correctly detect 86.7% of real disease instances and this will be quite significant for avoiding the cases of missing diagnosis in commercial poultry farms. At the standard 0.5 IoU detection threshold (mAP@0.5) the average precision was 0.971, indicating the localization of disease symptoms to be superior under common detection circumstances.

Measurement over the broader numbers of intersection-over-union thresholds of 0.5 to 0.95 (mAP@0.5-0.95), the model showed considerable results of 0.724, which shows that it has a reliable detection ability even with the more strenuous localization condition. All of these findings indicate that though the system exhibits unimaginable sensitivity in disease detection (high recall) and the strong performance in accurate localization (high mAP), it can be further optimized to lower the false positive rates with stronger training on the hard negative samples or due to the inclusion of multi-modal data inputs. This model is especially suited to use in the real world since the performance features allow it to be well-suited when complete screening of diseases is most important rather than ideal specificity.

## CONCLUSION

This behaviour has vowed to bring significant learning to real time monitoring of the poultry health by being accurate and efficient in diseased patch detection of photos of hens. Powerful pipeline to tackle the challenges of manually detecting the diseases on chicken farming was developed, which employed Google Colab in training, LabelImg in precise annotation, and Roboflow in dataset preparation.

Poultry disease classification works well on this system, and the same can be said of the balance between sensitivity (recall = 0.867) and localization accuracy (mAP@0.5 = 0.971). Although the model performs well to detect cases of true diseases especially in Fowl Pox and in healthy birds, false positives of Infectious Coryza and the recall of Newcastle Disease is still an issue. Such findings indicate a high potential of the system in a real-life application and a possibility of optimising it using specific training measures and multi-modal data-integrating to improve the accuracy and not jeopardise the sensitivity of detection.

This is significant progress because it becomes impractical to detect more than one infection at any time, and this is typical of the normal method of doing so, which is to either select or employ highly time consuming techniques, pertaining to a single infection. An effective solution to such an implementation gives not only a general increase in the comfort of animals, the possibility of minimizing economic losses and identifying dangerous diseases at an early stage also allows a fundamental change in the way such diseases are managed in the chicken farming.

In order to enhance the robustness of the model, future research might consider increasing the data to present a wider range of more varied photographs of chicken in various environments. The model is also incorporated to a chicken farm real-time monitoring system to serve the farmers with valuable information not forgetting solutions. The possibility to enhance the accuracy and practical utility of the model in the future can be achieved through studying the use of more sophisticated imaging techniques, including those of thermal or hyperspectral imaging. According to everything mentioned, this research establishes a basis on the extensive application of AI based solutions in the monitoring of poultry health which will augment production and sustainability of the industry.

ACKNOWLEDGMENT

We would like to acknowledge our most honored guide Dr. Vijayarani J., Assistant Professor, Department of Computer Science and Engineering, who has given us her valuable contribution, considerate motivation and outstanding preparation in the entire period of this project. Her minds eye suggestions, positive inputs and constant support have helped in moulding our work and us accomplishing our aims.

We are very glad also to the management of HINDUSTAN INSTITUTE OF TECHNOLOGY AND SCIENCE to give us the resources and facilities needed in the effort of carrying out our project work effectively. Their sponsorship is what has turned out to be the most critical factor in relieving the success of this venture.

Finally, we would like to express gratitude to our family, friends and colleagues who have been very inspirational, encouraging, and supportive all through this process. They have also supported us by lending a helping hand and believing in us and this has served us in beating the odds and meeting our goals.